\documentclass[a4paper,fleqn]{cas-dc}

\usepackage[numbers]{natbib}
\usepackage{algorithm}
\usepackage{diagbox}

\def\tsc#1{\csdef{#1}{\textsc{\lowercase{#1}}\xspace}}
\tsc{WGM}
\tsc{QE}

\begin{document}
\let\WriteBookmarks\relax
\def\floatpagepagefraction{1}
\def\textpagefraction{.001}
\let\printorcid\relax 

\shorttitle{3D Landmark Detection on Human Point Clouds: A Benchmark Dataset and Baseline}

\shortauthors{Fan Zhang et al.}

\title[mode = title]{3D Landmark Detection on Human Point Clouds: A Benchmark and A Dual Cascade Point Transformer Framework}  
\tnotemark[1,2]

\author[1,2]{Fan Zhang}
\cormark[1]
\ead{fanzhang@gatech.edu}

\author[3]{Shuyi Mao}
\ead{maosy6@lenovo.com}

\author[1]{Qing Li}
\ead{liqing@sztu.edu.cn}

\author[1]{Xiaojiang Peng}
\cormark[2] 
\ead{pengxiaojiang@sztu.edu.cn}

\address[1]{College of Big Data and Internet, Shenzhen Technology University, Shenzhen, China}
\address[2]{Department of Electrical and Computer Engineering, Georgia Institute of Technology, Shenzhen, China}
\address[3]{AI Lab, Lenovo Research, Shenzhen, China}

\cortext[1]{This work is done when Fan Zhang is an intern at Shenzhen Technology University.} 
\cortext[2]{Corresponding author.}

\begin{abstract}
3D landmark detection plays a pivotal role in various applications such as 3D registration, pose estimation, and virtual try-on. 
While considerable success has been achieved in 2D human landmark detection or pose estimation, there is a notable scarcity of reported works on landmark detection in unordered 3D point clouds. 
This paper introduces a novel challenge, namely 3D landmark detection on human point clouds, presenting two primary contributions. 
Firstly, we establish a comprehensive human point cloud dataset, named HPoint103, designed to support the 3D landmark detection community. 
This dataset comprises 103 human point clouds created with commercial software and actors, each manually annotated with 11 stable landmarks. 
Secondly, we propose a \textbf{D}ual \textbf{C}ascade \textbf{P}oint \textbf{T}ransformer (D-CPT) model for precise point-based landmark detection. 
D-CPT gradually refines the landmarks through cascade Transformer decoder layers across the entire point cloud stream, simultaneously enhancing landmark coordinates with a RefineNet over local regions.
Comparative evaluations with popular point-based methods on HPoint103 and the public dataset DHP19 demonstrate the dramatic outperformance of our D-CPT. 
Additionally, the integration of our RefineNet into existing methods consistently improves performance.
Code and data will be released soon.
\end{abstract}




\maketitle

\section{Introduction}
\label{intro}
In various upstream tasks that extensively involve landmark positions, such as 3D registration, pose estimation, and 3D virtual try-on, indispensable research focuses~\cite{carreira2016human,xiao2018simple} on accurately detecting landmark information.
Most existing tasks related to human landmark detection or pose estimation predominantly operate on 2D images, pinpointing the 2D or 3D human joints through direct regression or heatmap estimation. The success of this task is largely attributed to the implementation of deep convolutional neural networks (CNNs) and extensive structured image data \cite{carreira2016human,toshev2014deeppose,newell2016stacked,wei2016convolutional,xiao2018simple}. 
Contemporary approaches typically utilize CNN backbones for feature extraction from 2D images, though they often overlook landmarks with unclear spatial positions.
To tackle this limitation, one intuitive solution is to employ 3D point clouds for human landmark detection instead of 2D images.
However, to the best of our knowledge, the exploration of 3D landmark detection on human point clouds has been limited, a phenomenon we attribute to two primary challenges. 
Firstly, the collection of large-scale, high-quality human point clouds poses difficulties. Secondly, the unordered data structure complicates the adaptation of CNN-based landmark detection methods to point clouds.

\begin{figure}[t]
\centering
\includegraphics[width=\linewidth]{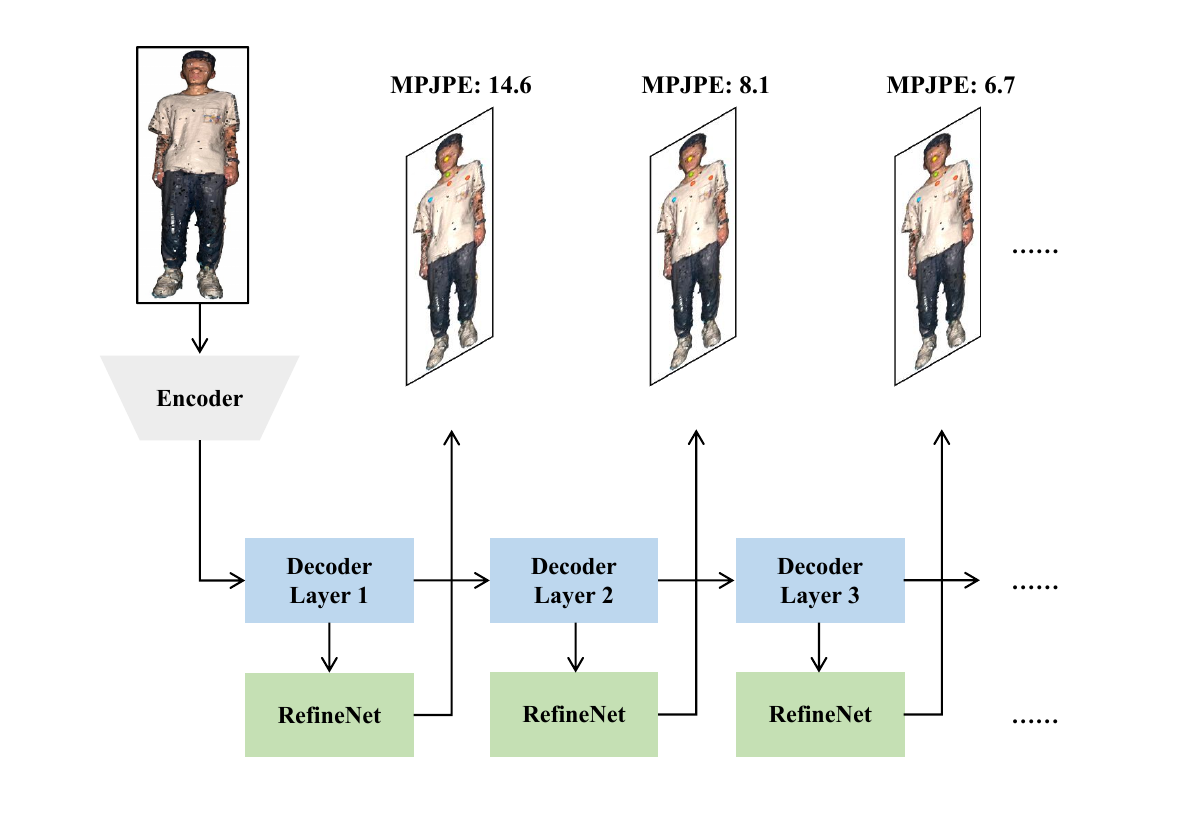}
\caption{Illustration of our dual cascade framework. The horizontal cascade process gradually refines results over the whole human point cloud model, while the vertical cascade process refines each landmark over local regions.}
\label{fig:1}
\end{figure}

In this paper, to fill the gap, we take the initiative to create a genuine and high-quality human point cloud dataset, termed as HPoint103. 
Employing a combination of a turntable and a camera, we capture high-resolution videos against a blue background.
Subsequently, we perform human matting to eliminate the background and input the video frames into commercial software for 3D point cloud generation.
The HPoint103 dataset comprises a total of 103 human point clouds, each ranging from 300k to 2,000k points, with each 3D point containing both coordinate and texture information. 
These point clouds are characterized by their richness and diversity in terms of human attire, genders, and postures. 
To enhance the dataset, we manually annotate each point cloud with 11 stable landmarks situated around the heads, shoulders, and knees.
\begin{figure*}[t]
\centering
\includegraphics[width=0.9\linewidth]{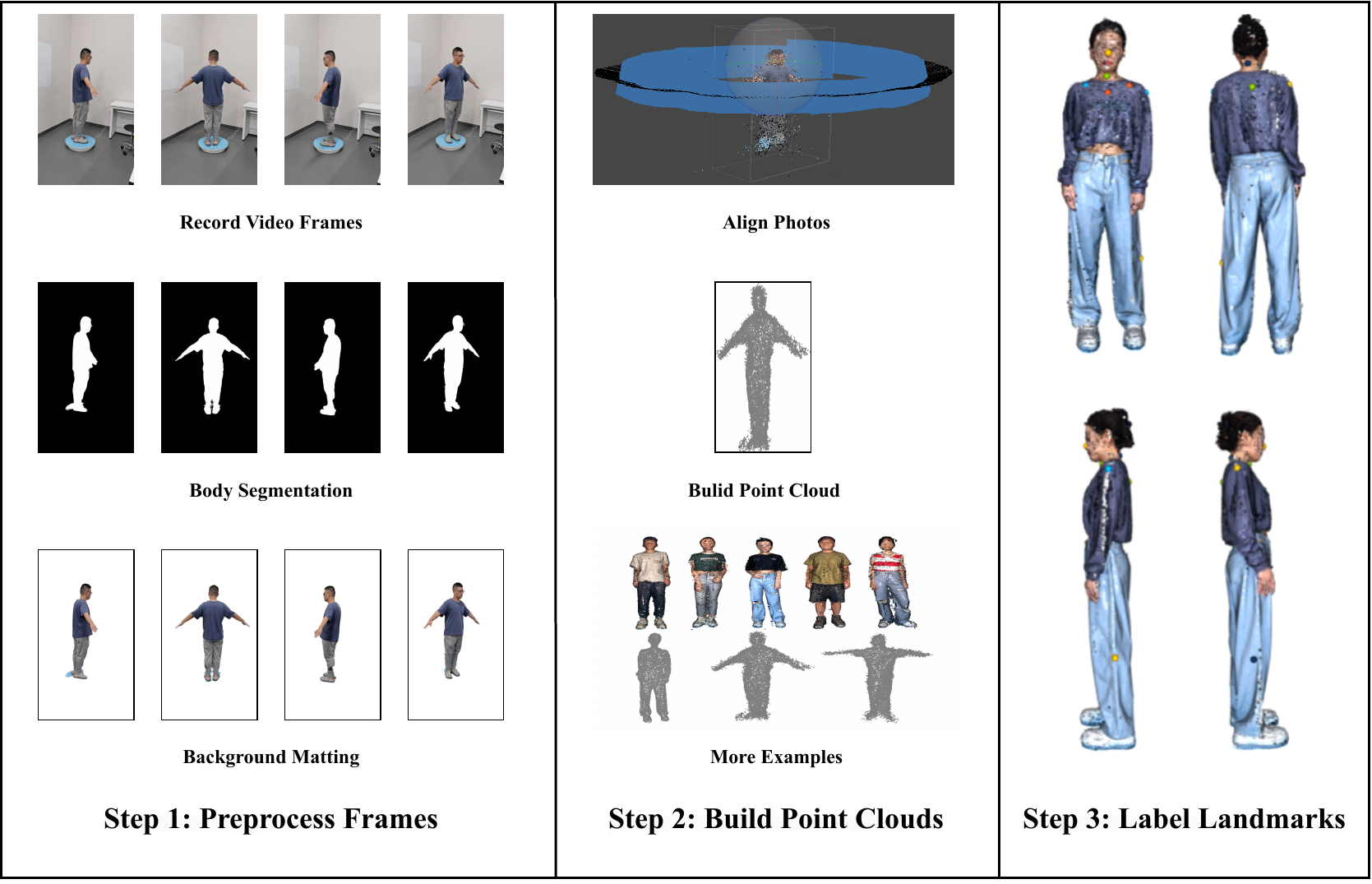}
\caption{The building steps of our HPoint103, include video recording, human matting, point cloud generation, and point annotation.}
\label{dataset}
\end{figure*}

In addition, to achieve precise landmark localization on point clouds, we introduce a straightforward yet novel framework named the Dual Cascade Point Transformer (D-CPT), depicted in Figure \ref{fig:1}. 
Leveraging the inherent point permutation invariance of Transformers, we employ them as decoders for landmark localization.
D-CPT comprises multiple Transformer decoder layers arranged in a dual cascade pipeline. 
The horizontal cascade process refines outcomes by stacking several decoders across the entire point cloud (\textit{i.e.}, CPT), constituting a multi-level supervision procedure during training. 
The vertical cascade process utilizes an additional Transformer decoder layer (\textit{i.e.}, RefineNet) that takes inputs from coarse predictions and enhances them within local point regions. 
Extensive experiments conducted on our HPoint103 dataset and DHP19 \cite{chen2022EPP,Calabrese_2019_CVPR_Workshops} dataset (a synthetic point cloud dataset, the sole relevant dataset identified) demonstrate the superiority of our method over popular point-based methods, both qualitatively and quantitatively. 
Furthermore, the integration of our RefineNet into existing methods consistently enhances performance.

In summary, this paper makes contributions in the following aspects:
\begin{itemize}
		\item \textbf{Problem Formulation.} We pioneer the exploration of 3D landmark detection on human point clouds, marking a crucial step for further investigations into upstream tasks such as 3D human pose estimation, 3D head swap, and 3D virtual try-on.
		\item \textbf{Real-world Dataset.} We establish a genuine and high-quality human point cloud dataset HPoint103, with each model meticulously annotated with 11 stable and easily discernible landmarks. The dataset will be made publicly available to support advancements in the field of 3D human landmark detection. 
		\item \textbf{Novel Methodology.} We intricately devise the Dual Cascade Point Transformer model, showcasing its remarkable superiority over existing popular methods for point cloud analysis. 
        \item \textbf{Comprehensive Experiments.} Through comprehensive experiments on HPoint103 and DHP19, we verify the superiority of our method over the existing popular point-cloud analysis methods. Additionally, the sub-module of this model holds promise for seamless integration into other methods, offering efficient performance improvements.
\end{itemize} 

\begin{figure*}[t]
\centering
\includegraphics[width=0.9\linewidth]{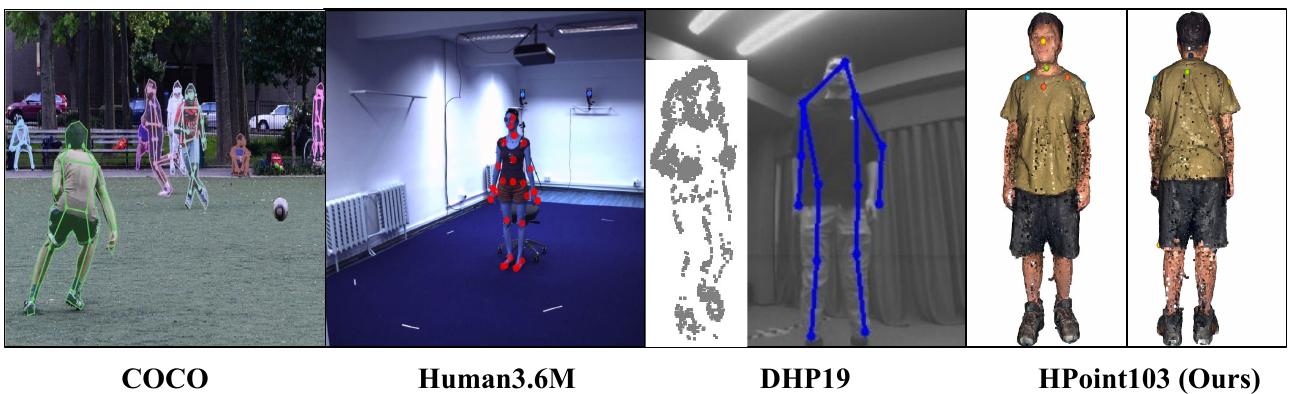}
\caption{Qualitive comparison between different human landmark datasets. The left part of DHP19 is the point cloud converted by frames, which is far more worse than ours.}
\label{fig:6}
\end{figure*}

\section{Related Work}
\label{related work}

\subsection{Human Landmark Detection}
Landmark detection is widely adopted among human pose estimation (HPE) tasks. Both 2D and 3D HPE approaches can be divided into regression-based \cite{carreira2016human,toshev2014deeppose} and heatmap-based methods \cite{newell2016stacked,wei2016convolutional,xiao2018simple}. 
The regression-based methods predict the coordinates of landmarks directly, while the heatmap-based methods predict heatmaps instead of coordinates and select the positions with maximum conﬁdence as the ﬁnal prediction. 
The former tends to be more concise and the latter is usually more precise. 
For 3D HPE, almost all methods are still based on 2D data such as single or multiple camera views and video frames. 
Even though the output target is 3D coordinates, these methods actually make 2D to 3D conversions, which is not as direct and efficient as our proposed point-to-point method. 

There are benchmark datasets such as Human3.6M \cite{h36m_pami}, MPI-INF-3DHP \cite{mono-3dhp2017}, and CMU Panoptic \cite{Joo_2015_ICCV,Joo_2017_TPAMI} for 3D HPE research. 
However, all of these datasets are made up of 2D images and are not suitable for our task. Therefore, in order to test the effectiveness of our method and to encourage more work to boost in this area, we build a brand new dataset called HPoint103, consisting of human point cloud data with stable landmark labels.

\begin{table}[t]
\centering
\resizebox{\linewidth}{!}{%
\begin{tabular}{c|c|c|c}
\toprule
Dataset   & Data Format & Landmark Format & Landmark Amount \\
\midrule
COCO      & RGB Image   & 2D              & 19              \\
Human3.6M  & RGB Image   & 2D/3D            & 32              \\
DHP19     & Event Frame & 2D/3D             & 13              \\
HPoint103 & Point Cloud & 3D              & 11             \\
\bottomrule
\end{tabular}%
}
\caption{Quantitative comparison between different human landmark datasets. HPoint103 consists of point cloud data with 3D landmarks.}
\label{tab:2}
\end{table}

\subsection{Point Cloud Analysis}
Rather than projecting 3D point clouds into 2D image planes \cite{su2015multi,li2016vehicle,chen2017multi,kanezaki2018rotationnet,lang2019pointpillars} or transforming irregular point clouds to regular voxels \cite{maturana2015voxnet,song2017semantic,zhou2018voxelnet}, deep networks that ingest point clouds directly can process 3D data in a more graceful and efficient manner. PointNet \cite{qi2017pointnet}, with pointwise MLPs and pooling layers utilized as permutation-invariant operators to aggregate features, is a pioneer in this series of works. PointNet++ \cite{qi2017pointnet++} goes one step further by adopting a hierarchical spatial structure to increase the sensitivity to local features. Another number of approaches combine point sets with graph structures. DGCNN \cite{wang2019dynamic} with dynamic graph convolutions on k-nearest neighbors (kNN) graphs is the most representative work. Some researchers \cite{guo2021pct,zhao2021point,park2022fast} have also begun to study how to transfer Transformer structure from 2D image understanding \cite{liu2021swin,dosovitskiy2020image,ramachandran2019stand} to 3D point clouds efficiently. 
For research on point clouds, current works mainly focus on object classification \cite{tian2023joint,zhang20233d,zhou2024tnpc}, semantic segmentation \cite{saglam2020boundary,park2023pcscnet,yang2023extracting,yang2023extracting,10319084}, object detection \cite{10313022}, and registration \cite{zaman2023cmdgat,wu2024surface,10313937,10313949}. It is still an open challenge about how they work on regression tasks like landmark detection.

\subsection{Transformers for Point Cloud Understanding}
According to \cite{zhao2020exploring}, self-attention mechanisms can be categorized into scalar dot-product attention and vector dot-product attention.  
Zhao et al.~\cite{zhao2021point} and Guo et al.~\cite{guo2021pct} have emerged as trailblazers in the application of attention mechanisms to point cloud understanding.
PCT~\cite{guo2021pct} introduces global attention directly onto the point cloud, akin to ViT~~\cite{dosovitskiy2020image}. 
In contrast, leveraging the vector attention theory proposed in SAN~\cite{zhao2020exploring}, Point Transformer~\cite{zhao2021point} conducts local attention between each point and its adjacent points, mitigating the aforementioned memory challenges.
Point Transformer~\cite{zhao2021point} has exhibited outstanding performance across various point cloud understanding tasks and has achieved state-of-the-art results in several competitive challenges.

\section{HPoint103}
\label{HPoint103}

\begin{figure*}[t]
\centering
\includegraphics[width=\linewidth]{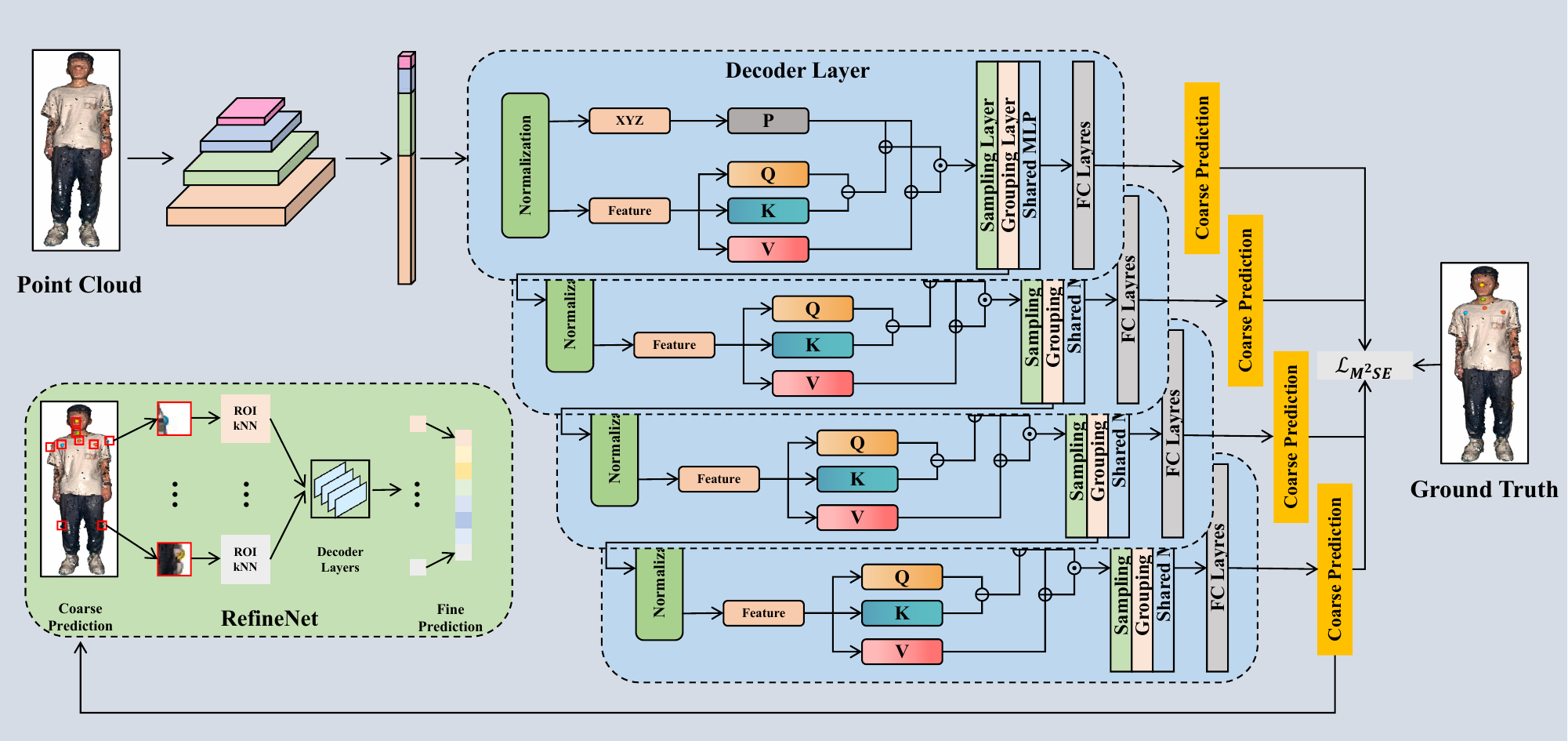}
\caption{The pipeline of our proposed network. The input point cloud is first encoded into the point-wise feature. The hierarchical point-wise feature after the cascade decoding process is then transformed into the coarse prediction and sent to the RefineNet. In the refinement process, each landmark is upsampled to $k$ points through kNN search in the region of interest. The input coarse prediction is then refined into the fine prediction.}
\label{fig:2}
\end{figure*}

\subsection{Dataset Collection and Annotation}
To establish a benchmark for the point cloud landmark detection task, we meticulously construct the HPoint103 dataset, as depicted in Figure \ref{dataset}.
The data collection involves placing the actor on a turntable, and rotating with it, while a stationary camera captures high-resolution videos, generating multi-view frames centered around the human subject.
To eliminate irrelevant background elements, we employ BackgroundMattingV2~\cite{BGMv2} with a pre-recorded background video.
The subsequent step involves constructing point clouds for each actor using Structure-from-Motion (SfM) and Multi-View Stereo (MVS) pipelines. 
While COLMAP \cite{Schonberger_2016_CVPR,schoenberger2016mvs} serves as an open-source toolbox for dense point cloud generation, its efficiency diminishes considerably when handling millions of points. 
Consequently, we opt for the commercial software Metashape for its expeditious processing. 
For each actor, we capture over 300 undistorted images from various angles, converting them into a point cloud. 
The point clouds exhibit varying point counts, ranging from 300k to 2,000k points, with each model point encompassing x-y-z coordinates, normals, and colors.

Following the point cloud collection, the subsequent phase involves landmark annotation.
We design a proprietary annotation tool utilizing Open3d, facilitating the annotation of 11 stable landmarks. Subsequently, two experts independently annotate each point cloud, and the averaged coordinates from their annotations are considered as the final annotations. 
As illustrated on the right side of Figure \ref{dataset}, the landmarks are strategically positioned around the heads, shoulders, and knees.
The rationale behind selecting these specific landmarks lies in their ability to effectively encapsulate geometric information pertaining to the human head. 
This choice is motivated by the utility of such information in prospective studies on tasks such as 3D head swapping and 3D virtual try-on.

\subsection{Dataset Comparison}
We conduct a comprehensive comparison between our HPoint103 and three benchmark datasets, as illustrated in Figure \ref{fig:6} and detailed in Table \ref{tab:2}. 
The selected benchmark datasets, namely COCO \cite{lin2014microsoft}, Human3.6M \cite{h36m_pami}, and DHP19 \cite{Calabrese_2019_CVPR_Workshops}, are renowned in the domains of 2D and 3D human pose estimation. 
These datasets primarily comprise 2D images or event-based frames captured from single or multi-view cameras, with 19, 32, and 13 annotated landmarks, respectively. 
COCO annotates landmarks with coordinates and an additional visible indicator, while DHP19 labels 2D landmarks and reconstructs sparse points using event cameras.
In contrast, HPoint103 stands out by eliminating unrelated backgrounds and generating pure human point clouds through the utilization of commercial 3D reconstruction software. Despite the potentially higher sample counts in the other datasets, HPoint103 distinguishes itself as the sole 3D landmark dataset derived from real-world scanned point clouds.

\section{Methodology}
\label{sec:method}

\subsection{Overview of D-CPT}
The pipeline of D-CPT is depicted in Figure \ref{fig:2}. 
The initial step involves feeding the input point cloud into a Multi-Layer Perceptron (MLP) acting as the encoder, transforming it into point-wise features. 
Two pivotal modules are intricately designed within the proposed network. 
The first module encompasses the cascade decoder layer, comprising a point Transformer decoder and a pooling layer. 
The performance of our network is significantly enhanced by the stacking of multiple cascade decoder layers.
The second module, \textit{i.e.}, a RefineNet, plays a crucial role in refining the localization of each individual landmark point within local regions.
The subsequent sections will delve into a detailed discussion of the design and functionality of each module.

\subsection{Transformers for 3D Landmark Detection} 
Motivated by Zhao et al. \cite{zhao2021point}, we introduce a novel local vector dot-product attention mechanism within the point Transformer decoder to consolidate local features for precise landmark localization. 
Diverging from conventional self-attention mechanisms found in natural language processing and image analysis, the vector attention mechanism applied to point clouds can be characterized as follows:
\begin{equation}
F(X)=\sum_{p_{i},f_{i} \epsilon X}^{}Softmax(Q-K+P)\odot (V+P),
\end{equation}
where $p_{i},f_{i}\epsilon X$  are sampled points and features from the point set. $\odot $ is defined as Hadamard Product. Then $Q$, $K$, and $V$ are defined as:   
\begin{equation}
  \begin{aligned}
  & x_{i}=FC_{1}(f_{i}),\\
  & Q=FC_{2}(x_{i}), \\
  & K=kNN(FC_{3}(x_{i})), \\
  & V=kNN(FC_{4}(x_{i})). 
  \end{aligned}
\end{equation}
In this approach, distinct fully connected layers are employed for $Q$, $K$, and $V$ to embed features for subsequent computations. 
Following the fully connected layers, kNN points are queried for $K$ and $V$ to compute the attention mechanism, thereby aggregating local features for subsequent spatial locations. 
It is important to note that the dot-product attention with $f_{i}$ is not computed directly. 
Instead, $f_{i}$ undergoes encoding into $x_{i}$ post the fully connected layers before being incorporated into the computation. 
The positional encoding is defined as:
\begin{equation}
P=FC_{5}(p_{i}-kNN(p_{i})).
\end{equation}
Conventional positional encoding schemes for sequences or image patches often rely on manually crafted values, typically derived from sine or cosine functions. 
However, considering that point clouds represent unordered sets of points inherently endowed with positional information, the coordinates themselves become naturally suited for positional encoding schemes. 
In this context, the subtraction relation of coordinates between a given point and the sampled k points is deemed an effective positional relation, and as such, we encode this relation as positional encoding.

It is crucial to highlight that the input point clouds can assume arbitrary orientations, posing a significant challenge in 3D point cloud location-related tasks. 
While some prior works overlook this challenge, primarily focusing on classification or segmentation tasks, where segmentation is essentially treated as a specialized form of classification, the arbitrary orientation issue is often disregarded. 
This neglect is caused by the robustness of the proposed networks in encoding features for classification. 
However, in regression tasks such as location detection, orientation becomes pivotal for accurate predictions. 
Leveraging positional encoding within the self-attention mechanism enables the network to incorporate position with orientation information, effectively addressing the arbitrary orientation problem without the necessity of introducing additional modules, such as the T-Net in PointNet \cite{qi2017pointnet} or the oriented bounding box (OBB) in Hand PointNet \cite{ge2018hand}.

\subsection{Dual Cascade Architecture}
\subsubsection{Horizontal Cascade Process}
Following PointNet++ \cite{qi2017pointnet++}, we adopt a hierarchical spatial structure as the pooling layer for sampling and grouping input points. 
The pooling layer comprises three pivotal components: the sampling layer, the grouping layer, and the shared Multi-Layer Perceptron (MLP) layer. 
In the sampling layer, we employ iterative farthest point sampling (FPS) to effectively downsample the number of points, while the grouping layer consolidates kNN points around the centroid. 
The shared MLP layer utilizes shared weights, facilitating the aggregation of local features from grouped points for subsequent utilization. 
Throughout each pooling process within the decoder layer, the points encompassed in the point set undergo sampling and grouping, ultimately reducing their count to $1/4$ of the original amount.

The primary role of the pooling layer is to efficiently reduce the number of points while grouping them together. 
This process allows the subsequent point Transformer decoder in the next layer to leverage the aggregated local features for self-attention computation. 
Specifically, we conduct a search for the 16 nearest neighbors for each sampled point, effectively compressing the input vector from $n \times d$ to $n' \times d'$ before feeding it into the subsequent decoder layer.

Following normalization, the input vectors are partitioned into coordinates and features for subsequent computations. 
It is crucial to note that normalization plays a pivotal role in achieving optimal prediction results and the absence of normalization may lead to a decline in accuracy. 
The output size of the point Transformer decoder is $n \times f$, and it is further fed into the pooling layer. 
The synergy between the point Transformer decoder and the pooling layer endows the decoder layer with the capability to extract hierarchical local features by computing local self-attention on different scales. 
The integration of cascade decoder layers significantly enhances the performance of our proposed framework. 
However, to mitigate the risk of overfitting, we strike a balance by setting $M = 4$. 
A subsequent fully connected layer is employed to regress the coordinates of the landmarks. Given the presence of $J$ landmarks, each with $x$, $y$, and $z$ coordinates, the output size is configured as $j \times 3$. During the training stage, we implement a multi-level supervision procedure employing the Mean Square Error (MSE) loss. 
The formulation of the multistage MSE loss $\mathcal{L}_{M^{2}SE}$ is as follows:
\begin{equation}
\mathcal{L}_{M^{2}SE} = \frac{1}{J}\sum_{k\epsilon S_{k}} \sum_{{i}\epsilon{J}} (y_{i}-p_{i,k}(x_i,\theta ))^{2},
\end{equation}
where $p_{i,k}(x_i,\theta )$ represents the prediction of the $i$-th landmark from the $k$-th stage $S_{k}$. $\mathcal{L}_{M^{2}SE}$ is applied to penalize the deviation between the prediction of each decoder layer and the ground truth, which is different from other hierarchical models like PointNet++ and Point Transformer. 

\begin{table*}[t]
\centering
\resizebox{0.8\linewidth}{!}{%
\begin{tabular}{l|ccc|ccc}
\toprule
 & \multicolumn{3}{c|}{Sparse}  & \multicolumn{3}{c}{Dense}                       \\ 
\multirow{-2}{*}{Method} & Params (M)   & FLOPs (G)       & MPJPE (3D) $\downarrow$   & Params (M)   & FLOPs (G)     & MPJPE (3D) $\downarrow$         \\ \midrule
PointNet                  & 3.47                        & 3.59                        & 15.12          & 3.47                        & 3.59                          & 16.24          \\ 
DGCNN                     & 25.47                       & 135.58                      & 9.80          & 25.47                       & 135.58                        & 43.94          \\ 
PointNet++ (MSG)          & 1.75                        & 8.14                        & 10.46          & 1.75                        & 8.14                          & 16.79          \\ 
PointNet++ (SSG)          & 1.47                        & 1.74                        & 10.73          & 1.47                        & 1.74                          & 11.62          \\ 
Point Transformer                & 9.58 & 147.53 & 7.40          & 9.58 &  590.13 & 7.71          \\ 

CPT (Ours)              & 0.99 & 3.23 & \textbf{6.73}          & 9.64 &  590.13 & \textbf{7.12}          \\ \bottomrule
\end{tabular}%
}
\caption{Quantitative comparisons on different scales of Hpoint103. CPT achieves the best performance on both scales.}
\label{tab1}
\end{table*}

\begin{table*}[h]
\centering
\resizebox{0.8\linewidth}{!}{%
\begin{tabular}{l|l|cccc}
\toprule
Input                           & Method            & Params (M) & FLOPs (G) & MPJPE (2D) $\downarrow$   & MPJPE (3D) $\downarrow$    \\
\midrule
\multirow{2}{*}{2D Frames}      
                                & LeViT-128S        & 7.87       & 0.40      & 7.68          & 87.79          \\
                                & DHP19             & 0.22       & 7.02      & 7.67          & 87.90          \\
\midrule
\multirow{4}{*}{3D Point Clouds} & PointNet          & 4.46       & 2.38      & 7.29          & 82.46          \\
                                & DGCNN             & 4.51       & 9.82      & 6.83          & 77.32          \\
                                & Point Transformer & 3.65       & 10.06     & 6.46          & 73.37          \\
                                & CPT(Ours)         & 4.36       & 10.10     & \textbf{6.33} & \textbf{71.81}\\
\bottomrule
\end{tabular}%
}
\caption{Quantitative comparisons on DHP19. CPT outperforms the existing point-based methods on both 2D and 3D tasks. What's more, our method even beats some CNN-based methods as well.}
\label{DHP19}
\end{table*}

\subsubsection{Vertical Cascade Process}
Utilizing fully connected layers following the pooling layer, the network inherently possesses the capability to discern the locations of landmarks. 
In a pursuit to enhance prediction accuracy, we introduce a RefineNet designed to refine locations and mitigate errors. 
This process entails a global-to-local cascade, where kNN points surrounding each landmark within the region of interest are systematically sought to refine the predictions.

Upon predicting the coarse location of each landmark, a KD-Tree is generated for the corresponding landmark. 
Subsequently, we leverage the KD-Tree to conduct a search for kNN points surrounding the landmark within local regions, forming a new point cloud with a size of $k \times d$. 
Given the time-consuming nature of kNN search in relatively large point clouds, the creation of a KD-Tree is imperative. 
This not only expedites the process but also reduces the computational complexity from $O(n)$ to $O(\log n)$. 
Each landmark is associated with its KD-Tree, resulting in a total of $J$ KD-Trees. Harnessing the capabilities of the regression-based network, the KD-Tree is seamlessly integrated with the decoder layers, collectively forming a RefineNet. 
Subsequently, the new point cloud undergoes refinement through stacked decoder layers to improve the precision of predictions.

The rationale behind the RefineNet is elegantly simple yet highly effective: predicting the locations of landmarks in a smaller point cloud is inherently easier than in a larger point cloud. 
This stems from the inductive bias that the exact coordinates of landmarks typically do not deviate significantly from those predicted in the initial stage. 
As a result, we leverage the coarse prediction as the starting point and search for $k$ points around it to refine the prediction further, yielding a substantially more accurate outcome than the coarse prediction alone. 
Notably, the choice of the parameter $k$ is pivotal to the final prediction, and inappropriate values may lead to performance degradation. 
The selection of optimal $k$ values will be thoroughly discussed in the ablation section.

\section{Experiment}
\label{sec:experiments}

\subsection{Metrics and Implementation Details}
The Mean Per Joint Position Error (MPJPE), a widely employed metric in human pose estimation, is adopted for evaluating the predictions. MPJPE is defined as:
\begin{equation}
MPJPE=\frac{1}{J} \sum_{{i}\epsilon{J}} \left \| pred_{i}-gt_{i}   \right \|_2,
\end{equation}
where $pred_{i}$ and $gt_{i}$ are respectively the prediction and ground truth of the $i$-th landmark. In total, there are $J$ landmarks for each individual.

We employ the Adam optimizer with default parameters during training. The initial learning rate is set to $0.001$, and the training process is stopped after $100$ epochs. For the point Transformer decoder, we choose $16$ nearest neighbors for each point. During the refinement stage, when the number of sampled points is $4096$ and $16384$, we set $k$ to $1024$ and $4096$ respectively. The RefineNet comprises $4$ decoder layers. Dataset splitting involves allocating $90\%$ for training and $10\%$ for inference. All experiments are conducted using PyTorch on NVIDIA Tesla A100 GPUs.

\begin{table}[t]
\centering
\begin{tabular}{l|c}
\toprule
{\color[HTML]{000000} Method} & {\color[HTML]{000000} MPJPE (3D) $\downarrow$} \\
\midrule
{\color[HTML]{000000} PointNet w/o RefineNet} & {\color[HTML]{000000} 15.12} \\
{\color[HTML]{000000} PointNet w RefineNet} & {\color[HTML]{000000} \textbf{13.84}} \\
\midrule
{\color[HTML]{000000} DGCNN w/o RefineNet} & {\color[HTML]{000000} 9.80} \\
{\color[HTML]{000000} DGCNN w RefineNet} & {\color[HTML]{000000} \textbf{9.32}} \\
\midrule
{\color[HTML]{000000} PointNet++ (MSG) w/o RefineNet} & {\color[HTML]{000000} 10.46} \\
{\color[HTML]{000000} PointNet++ (MSG) w RefineNet} & {\color[HTML]{000000} \textbf{10.20}} \\
\midrule
{\color[HTML]{000000} PointNet++ (SSG) w/o RefineNet} & {\color[HTML]{000000} 10.73} \\
{\color[HTML]{000000} PointNet++ (SSG) w RefineNet} & {\color[HTML]{000000} \textbf{9.98}}\\
\midrule
{\color[HTML]{000000} Point Transformer w/o RefineNet} & {\color[HTML]{000000} 7.40} \\
{\color[HTML]{000000} Point Transformer w RefineNet} & {\color[HTML]{000000} \textbf{6.91}}\\
\midrule
{\color[HTML]{000000} CPT (Ours)} & {\color[HTML]{000000} 6.73} \\
{\color[HTML]{000000} D-CPT (Ours)} & {\color[HTML]{000000} \textbf{6.44}} \\
\bottomrule
\end{tabular}%
\caption{Extend the RefineNet to point-based approaches on HPoint103. RefineNet is not designed specifically for CPT, but a plug-and-play module. It helps refine the prediction over local regions and thus achieve smaller MPJPE.}
\label{extend refinement}
\end{table}

\subsection{Comparison with Baselines}
To the best of our knowledge, no prior work has addressed the task of detecting human landmarks by taking point clouds as input. Mainstream approaches predominantly focus on landmark localization from 2D images. While some previous works, such as Hand PointNet \cite{ge2018hand} and \cite{o20203d}, have employed point-based networks for detecting landmarks on hands and ears, the point sets associated with these body parts are considerably smaller compared to the human form. Consequently, detecting landmarks in such large and sparse point clouds presents a formidable challenge.

In our comparative experiments, we evaluate a set of widely recognized point-based networks, including PointNet \cite{qi2017pointnet}, DGCNN \cite{wang2019dynamic}, PointNet++ \cite{qi2017pointnet++}, and Point Transformer \cite{zhao2021point}, which are originally designed for tasks other than regression. We leverage each of these networks as a backbone architecture, modifying the last two linear layers to perform landmark coordinates regression. Given the impact of the number of input sampled points on network performance, we consider two scales: Sparse (points size $4096 \times 3$) and Dense (points size $16384 \times 3$).

\begin{table}[t]
\centering
\begin{tabular}{c|c|c|c}
\toprule
\diagbox  {Dimension}{Points} & 4096 & 16384 & 30000 \\
\midrule
3 & 7.40 & 7.12 & 21.04 \\
6 & 8.26 & 7.58 & 19.86 \\
10 & 1.38 & 1.37 & 2.53\\
\bottomrule
\end{tabular}%
 \caption{MPJPE results on different sampled points and feature dimensions of HPoint103.}
\label{size}
\end{table}

\begin{figure}[t]
\centering
\includegraphics[width=0.8\linewidth]{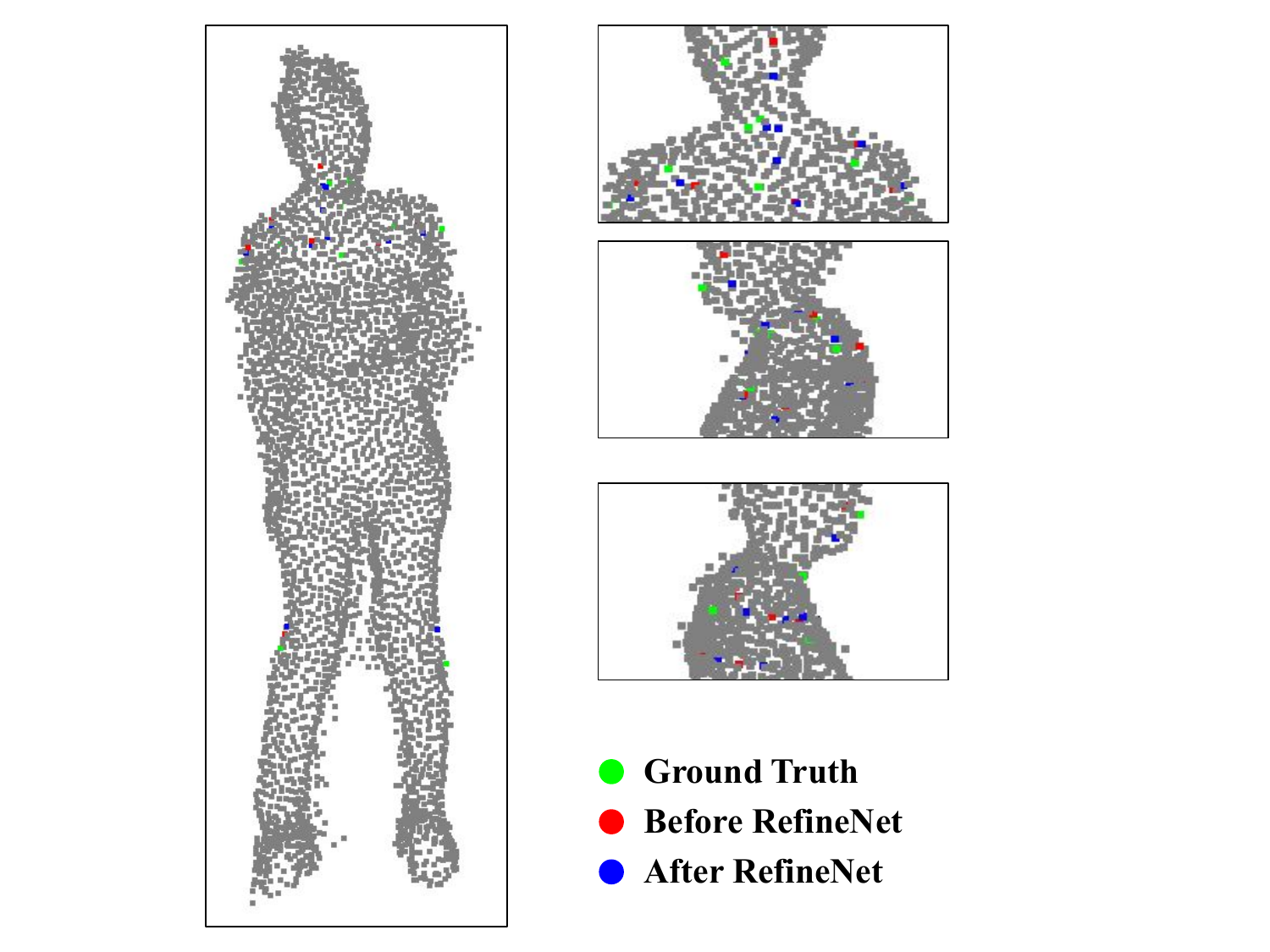}
\caption{Visualization of predicting locations before and after RefineNet. }
\label{fig:4}
\end{figure}

The experimental results on HPoint103, as presented in Table \ref{tab1}, demonstrate the superior performance of our method compared to the baseline approaches. Notably, we exclude reporting the performance of D-CPT at this point, as we consider RefineNet to be a sub-module, and it would be unfair to compare D-CPT with other point-based methods without considering RefineNet. A detailed analysis of the effectiveness of RefineNet will be provided in the subsequent sections.

In the Sparse scale, our approach attains the lowest error while utilizing the fewest parameters. When transitioning to the Dense scale, Transformer-based methods experience a significant increase in FLOPs due to channel-wise attention operations with dense points. Despite this, our approach outperforms others in this setting as well.

\subsection{Comparison on Public Dataset}
\label{cmpdhp19}
\begin{figure*}[t]
\centering
\includegraphics[width=0.9\linewidth]{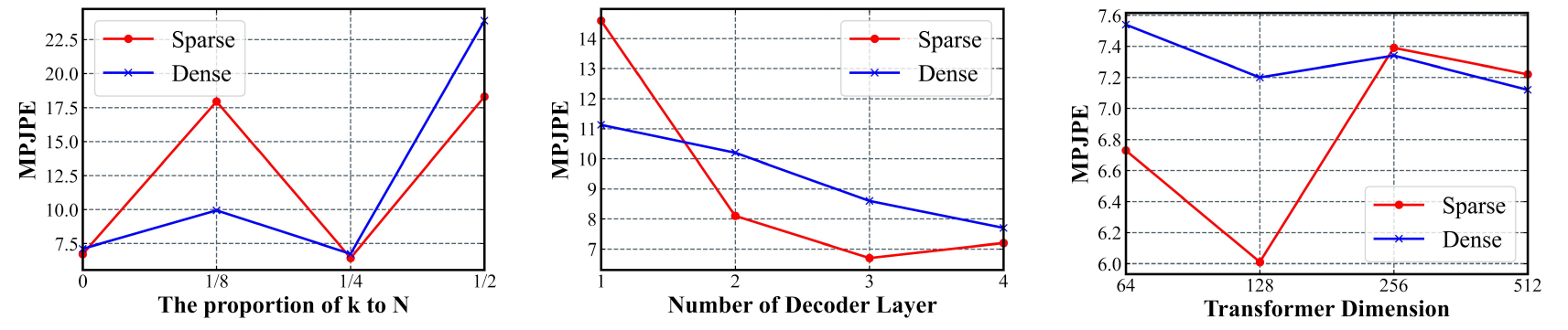}
\caption{Ablations on k value of RefineNet, depths of CPT, and widths of CPT.}
\label{ablation}
\end{figure*}

To assess the robustness of our method, we conduct an evaluation on the DHP19 dataset, a benchmark dataset focused on human pose estimation. Notably, EventPointPose \cite{chen2022EPP} recently proposes a technique to convert the 2D frames in DHP19 into 3D point clouds, facilitating their utilization for the 3D landmark detection task. This dataset is derived from high-quality recordings captured by four synchronized $346\times260$ pixel Dynamic Vision Sensors (DVS) cameras, combined with marker positions in 3D space obtained from the Vicon motion capture system. It encompasses 3D positions for a total of 17 subjects, each engaged in 33 distinct movements. The original dataset split designates S1-S12 for training and S13-S17 for testing, a configuration we adhere to in our validation. 

In our comparative analysis, we include two CNN-based methods, MobileHP-S \cite{choi2021mobilehumanpose} and DHP19 \cite{Calabrese_2019_CVPR_Workshops}. The results, showcased in Table \ref{DHP19}, underscore that our method outperforms all previous methods despite employing relatively compact model sizes. This superiority is not limited to point-based methods; even some CNN-based methods with comparable parameter counts fall short of achieving comparable performance. Furthermore, given the diverse human poses within the DHP19 dataset, our method exhibits robustness and invariance to rotation and rigid transformations.

\subsection{Evaluation on RefineNet}
\label{refine}
The RefineNet, serving as a versatile plug-and-play module, finds utility not only in our D-CPT but also seamlessly integrates into various point-based models. By taking the coarse prediction as input, the RefineNet module can be effortlessly extended to enhance the accuracy of other approaches. The evaluation results of our RefineNet module are presented in Table \ref{extend refinement}.

The experiments unequivocally demonstrate the efficacy of the RefineNet module in refining landmark locations. When combined with our D-CPT, the RefineNet contributes to achieving state-of-the-art performance. Notably, the performance of the RefineNet module is contingent on the accuracy of the coarse prediction at the first stage and the choice of different $k$ values. Figure \ref{fig:4} visually depicts prediction results before and after utilizing RefineNet. Both the output size of the decoder layer and the RefineNet are configured as $j \times 3$, presenting an advantage that heatmap-based methods cannot attain. This approach not only mirrors the high performance of heatmap-based methods but also embodies the simplicity and elegance characteristic of regression-based methods.

\subsection{Ablation Study}
\label{4.4}

\subsubsection{Number of points and feature dimension}
\label{scale}
We conduct a comprehensive evaluation of our approach on HPoint103 with varying sampled points and feature dimensions. To ensure a fair comparison, we employ CPT without the RefineNet module as the network and assess its performance across different input scales. The evaluation spans three primary scales.

Table \ref{size} elucidates that the prediction error initially increases and then decreases with the augmentation of input points. This phenomenon arises due to the exponential growth in computational cost associated with the vector dot-product attention mechanism as the number of input points escalates. Contrary to intuition, more points do not consistently yield better results, as selecting the right point from a dense set proves more challenging than from a sparse set. Consequently, when the number of points reaches $30k$, the accuracy is inferior to that of $16k$.

Moreover, in the context of feature dimensions, a dimension of $3$ implies that only the first $x$, $y$, and $z$ dimensions are input into the network. Conversely, a dimension of $6$ includes $nx$, $ny$, and $nz$, while $10$ encompasses the entire feature dimension, including $r$, $g$, and $b$ color information. As the feature dimension increases, more semantic information is conveyed to the network. Remarkably, the three dimensions $nx$, $ny$, and $nz$ in the middle may be considered as noise information, offering minimal contribution to the prediction result. The inclusion of position and color information, however, significantly enhances prediction accuracy.

\subsubsection{$K$ values of RefineNet}
\label{kvalues}
The efficacy of the proposed RefineNet is contingent on the selection of different $k$ values, as illustrated in the left segment of Figure \ref{ablation}. Inappropriate $k$ values may lead to a degradation in performance. When $k$ is too small, the network might fail to capture sufficient semantic information, while an excessively large $k$ can impede the refinement process. Based on empirical observations, we set $k$ to $1024$ for $4096$ sampled points and $4096$ for $16384$ sampled points. As a general guideline, choosing $k$ as $1/4$ of the number of input points has proven to yield optimal results.

\subsubsection{Depths of CPT}
We leverage cascade decoder layers to enhance the precision of our predictions. Generally, the performance improves with the stacking of multiple decoder layers. However, an excessive number of decoder layers may lead to overfitting and an increase in computational cost. The impact of varying numbers of decoders is depicted in the middle of Figure \ref{ablation}. Notably, on the sparse scale, we find that $3$ cascade decoder layers yield optimal results, while on the dense scale, the best performance is achieved with $4$ cascade decoder layers.

\subsubsection{Widths of CPT}
The widths of the CPT model play a crucial role in influencing prediction accuracy. To assess the impact of different embedding dimensions in the point Transformer decoder, we conduct width evaluation, as illustrated in the right part of Figure \ref{ablation}. The results reveal that, for the sparse and dense scales, dimensions of $128$ and $512$ respectively achieve optimal accuracy. Interestingly, it emerges that employing a wider decoder architecture is unnecessary when the input scale is not too large. This finding diverges from the requirements of Transformers in 2D image understanding scenarios. The ablation study on widths corroborates the results obtained from the analysis of depths – indicating that for relatively modest input sizes, adopting more complex structures (\textit{i.e.}, deeper and wider) is not imperative.

\section{Conclusion}
\label{sec:conclusion}
In this paper, we propose a novel approach for 3D human landmark detection, offering a fresh perspective of taking point clouds instead of 2D images. Our contribution includes the introduction of a new benchmark dataset, HPoint103, specifically designed for 3D human landmark detection. Moreover, we present a pioneering framework for detecting 3D landmarks on human point clouds, with the potential for extending the vertical cascade module, RefineNet, to augment other approaches. Experimental results demonstrate a substantial performance improvement over baseline methods, showcasing the efficacy of RefineNet in enhancing the overall model accuracy. In future work, we plan to expand the HPoint103 dataset and provide additional forms of annotations, such as instance annotations.

However, our approach has limitations, including suboptimal latency and an increase in model size as the input number of points rises. The channel-wise attention mechanism introduces a significant computational cost, necessitating a tradeoff between accuracy and inference speed. Despite these challenges, our current progress marks an initial exploration into this field, with ongoing research needed. We anticipate that the introduction of HPoint103 will stimulate further studies in the domain, and with continued efforts, these challenges can be addressed in future advancements.

\bibliographystyle{cas-model2-names}
\bibliography{cas-refs}

\end{document}